\documentclass[11pt,a4paper]{article}
\usepackage[hyperref]{ranlp2023}
\usepackage{times}
\usepackage{latexsym}

\usepackage{microtype}

\usepackage{inconsolata}
\usepackage[nolist]{acronym}
\usepackage{graphicx}
\usepackage{multirow}
\usepackage{subcaption}
\usepackage{adjustbox,booktabs}
\usepackage{dblfloatfix} 
\usepackage{xcolor}
\usepackage{booktabs}

\newcommand{\midsepremove}{\aboverulesep = 0mm \belowrulesep = 0mm}
\midsepremove
\newcommand{\midsepdefault}{\aboverulesep = 0.605mm \belowrulesep = 0.984mm}
\midsepdefault

\usepackage{array}
\newcolumntype{L}[1]{>{\raggedright\let\newline\\\arraybackslash\hspace{0pt}}m{#1}}
\newcolumntype{C}[1]{>{\centering\let\newline\\\arraybackslash\hspace{0pt}}m{#1}}
\newcolumntype{R}[1]{>{\raggedleft\let\newline\\\arraybackslash\hspace{0pt}}m{#1}}

\aclfinalcopy 

\title{Exploring the Landscape of Natural Language Processing Research}

\author{Tim Schopf, Karim Arabi, and Florian Matthes \\
        Technical University of Munich, Department of Computer Science, Germany \\
        \texttt{\{tim.schopf,karim.arabi,matthes\}@tum.de}}

\begin{acronym}
\acro{nlp}[NLP]{natural language processing}
\acro{lm}[LM]{Language Model}
\acro{plm}[PLM]{Pretrained Language Model}
\acroplural{plm}[PLMs]{Pretrained Language Models}
\acro{kg}[KG]{knowledge graph}
\acroplural{kg}[KGs]{knowledge graphs}
\acro{p}[P]{Precision}
\acro{r}[R]{Recall}
\acro{mag}[MAG]{Microsoft Academic Graph}
\acro{fos}[FoS]{field of study}
\acroplural{fos}[FoS]{fields of study}
\end{acronym}

\date{}

\begin{document}
\maketitle
\begin{abstract}
As an efficient approach to understand, generate, and process natural language texts, research in natural language processing (NLP) has exhibited a rapid spread and wide adoption in recent years. Given the increasing research work in this area, several NLP-related approaches have been surveyed in the research community. However, a comprehensive study that categorizes established topics, identifies trends, and outlines areas for future research remains absent. Contributing to closing this gap, we have systematically classified and analyzed research papers in the ACL Anthology. As a result, we present a structured overview of the research landscape, provide a taxonomy of fields of study in NLP, analyze recent developments in NLP, summarize our findings, and highlight directions for future work. \footnote{Code available: \href{https://github.com/sebischair/Exploring-NLP-Research}{https://github.com/sebischair/Exploring-NLP-Research}}
\end{abstract}

\section{Introduction}

Natural language is a fundamental aspect of human communication and inherent to human utterances and information sharing. Accordingly, most human-generated digital data are composed in natural language. Given the ever-increasing amount and importance of digital data, it is not surprising that computational linguists have started developing ideas on enabling machines to understand, generate, and process natural language since the 1950s \cite{hutchins-1999-retrospect}. 

\begin{figure}[ht!]
    \centering
    \includegraphics[width=\columnwidth]{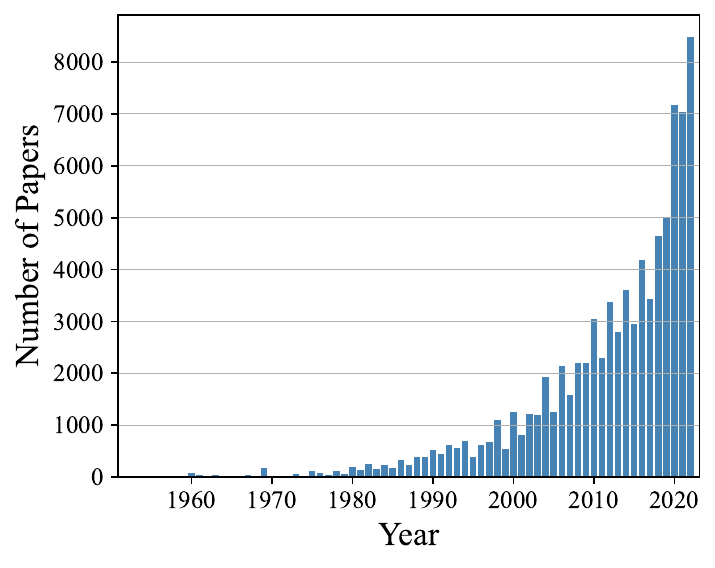}
    \caption{Distribution of the number of papers per year in the ACL Anthology from 1952 to 2022.}
    \label{fig:acl-anthology-number-of-papers}
\end{figure}

More recently, the introduction of the transformer model \cite{NIPS2017_3f5ee243} and pretrained language models \cite{Radford2018ImprovingLU,devlin-etal-2019-bert} have sparked increasing interest in \ac{nlp}. Submissions on various \ac{nlp} topics and applications are being published in a growing number of journals and conferences, such as TACL, ACL, and EMNLP, as well as in several smaller workshops that focus on specific areas. Thereby, the ACL Anthology\footnote{\href{https://aclanthology.org}{https://aclanthology.org}} as a repository for publications from many major \ac{nlp} journals, conferences, and workshops emerges as an important tool for researchers. As of January 2023, it provides access to over 80,000 articles published since 1952. Figure \ref{fig:acl-anthology-number-of-papers} shows the distribution of publications in the ACL Anthology over the 50-year observation period.

\begin{figure*}[ht!]
    \centering
    \includegraphics[width=\textwidth]{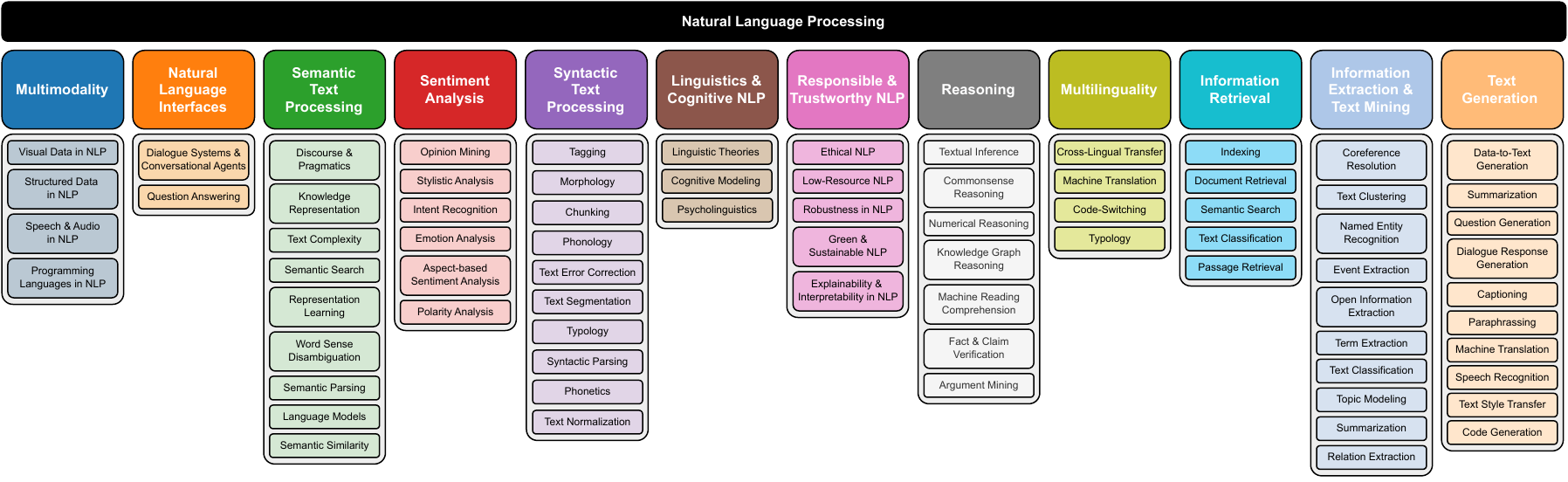}
    \caption{Taxonomy of fields of study in \ac{nlp}. Appendix \ref{sec:fos-descriptions} includes more detailed descriptions of the fields of study.}
    \label{fig:nlp-taxonomy}
\end{figure*}

Accompanying the increase in publications, there has also been a growth in the number of different \acp{fos} that have been researched within the \ac{nlp} domain. \ac{fos} are academic disciplines and concepts that usually consist of (but are not limited to) tasks or techniques \cite{shen-etal-2018-web}. Given the rapid developments in \ac{nlp} research, obtaining an overview of the domain and maintaining it is difficult. As such, collecting insights, consolidating existing results, and presenting a structured overview of the field is important. However, to the best of our knowledge, no studies exist yet that offer an overview of the entire landscape of \ac{nlp} research. To bridge this gap, we performed a comprehensive study to analyze all research performed in this area by classifying established topics, identifying trends, and outlining areas for future research. Our three main contributions are as follows:

\begin{itemize}
    \item We provide an extensive taxonomy of \acp{fos} in \ac{nlp} research shown in Figure \ref{fig:nlp-taxonomy}.
    \item We systematically classify research papers included in the ACL Anthology and report findings on the development of \acp{fos} in \ac{nlp}.
    \item We identify trends in \ac{nlp} research and highlight directions for future work.
\end{itemize}

Our study highlights the development and current state of \ac{nlp} research. Although we cannot fully cover all relevant work on this topic, we aim to provide a representative overview that can serve as a starting point for both \ac{nlp} scholars and practitioners. In addition, our analysis can assist the research community in bridging existing gaps and exploring various \acp{fos} in \ac{nlp}.

\section{Related Work}
\label{sec:related-work}
Related literature that considers various different \acp{fos} in \ac{nlp} is relatively scarce. Most studies focus only on a particular \ac{fos} or sub-field of \ac{nlp} research. 

For example, related studies focus on knowledge graphs in \ac{nlp} \cite{schneider-etal-2022-decade}, explainability in \ac{nlp} \cite{danilevsky-etal-2020-survey}, ethics and biases in \ac{nlp} \cite{suster-etal-2017-short,blodgett-etal-2020-language}, question answering \cite{liu-etal-2022-challenges}, or knowledge representations in language models \cite{safavi-koutra-2021-relational}. 

Studies that analyze \ac{nlp} research based on the entire ACL Anthology focus on citation analyses \cite{mohammad-2020-examining,rungta-etal-2022-geographic} or visualizations of venues, authors, and n-grams and keywords extracted from publications \cite{mohammad-2020-nlp-scholar,10.1007/978-3-030-45442-5_61}. 

\citet{anderson-etal-2012-towards} apply topic modeling to identify different epochs in the ACL’s history.

Various books categorize different \acp{fos} in \ac{nlp}, focusing on detailed explanations for each of these categories \cite{Allen1995-ALLNLU,10.5555/311445,UBMA_285413791,eisenstein2019introduction,tunstall2022natural}.



\section{Research Questions}
\label{subsec:research-questions}
The goal of our study is an extensive analysis of research performed in \ac{nlp} by classifying established topics, identifying trends, and outlining areas for future research. These objectives are reflected in our research questions (RQs) presented as follows:

\paragraph{RQ1:} \textit{What are the different \acp{fos} investigated in \ac{nlp} research?} 

Although most \acp{fos} in \ac{nlp} are well-known and defined, there currently exists no commonly used taxonomy or categorization scheme that attempts to collect and structure these \acp{fos} in a consistent and understandable format. Therefore, getting an overview of the entire field of \ac{nlp} research is difficult, especially for students and early career researchers. While there are lists of \ac{nlp} topics in conferences and textbooks, they tend to vary considerably and are often either too broad or too specialized. To classify and analyze developments in \ac{nlp}, we need a taxonomy that encompasses a wide range of different \acp{fos} in \ac{nlp}. Although this taxonomy may not include all possible \ac{nlp} concepts, it needs to cover a wide range of the most popular \acp{fos}, whereby missing \acp{fos} may be considered as subtopics of the included \acp{fos}. This taxonomy serves as an overarching classification scheme in which \ac{nlp} publications can be classified according to at least one of the included \acp{fos}, even if they do not directly address one of the \acp{fos}, but only subtopics thereof.

\paragraph{RQ2:} \textit{How to classify research publications according to the identified \acp{fos} in \ac{nlp}?} 

Classifying publications according to the identified \acp{fos} in \ac{nlp} is very tedious and time-consuming. Especially with a large number of \acp{fos} and publications, a manual approach is very costly. Therefore, we need an approach that can automatically classify publications according to the different \acp{fos} in \ac{nlp}. 

\paragraph{RQ3:} \textit{What are the characteristics and developments over time of the research literature in \ac{nlp}?} 

To understand past developments in \ac{nlp} research, we examine the evolution of popular \acp{fos} over time. This will allow a better understanding of current developments and help contextualize them. 

\paragraph{RQ4:} \textit{What are the current trends and directions of future work in \ac{nlp} research?} 

Analyzing the classified research publications allows us to identify current research trends and gaps and predict possible future developments in \ac{nlp} research.

\section{Classification \& Analysis}
\label{sec:results}

In this section, we report the approaches and results of the data classification and analysis. It is structured according to the formulated RQs.
\midsepremove
\begin{table*}[!h]
\renewcommand{\arraystretch}{0.71}
\begin{tabular}{@{}p{2.9cm}C{1.3cm}l|p{2.9cm}C{1.3cm}l@{}}
\toprule [1.5pt]
\small \textbf{Field of Study} & \small \textbf{\# Papers} & \small \textbf{Representative Papers} & \small \textbf{Field of Study} & \small \textbf{\# Papers} & \small \textbf{Representative Papers} \\  \midrule 
\multirow{2}{*}{\small Machine Translation} & \multirow{2}{*}{\small 12,922} & \multirow{2}{*}{\begin{tabular}[c]{@{}l@{}}\scriptsize \citet{liu-etal-2020-multilingual-denoising},  \\ \scriptsize \citet{goyal-etal-2022-flores}\end{tabular}} & \multirow{2}{*}{\small Visual Data in NLP} & \multirow{2}{*}{\small 2,401} & \multirow{2}{*}{\begin{tabular}[c]{@{}l@{}}\scriptsize \citet{tan-bansal-2019-lxmert}, \\ \scriptsize \citet{xu-etal-2021-layoutlmv2}\end{tabular}} \\
 &  &  &  &  &  \\ \midrule
\multirow{2}{*}{\small Language Models} & \multirow{2}{*}{\small 11,005} & \multirow{2}{*}{\begin{tabular}[c]{@{}l@{}}\scriptsize \citet{devlin-etal-2019-bert},\\ \scriptsize \citet{NEURIPS2022_b1efde53} \end{tabular}} & \multirow{2}{*}{\small Ethical NLP} & \multirow{2}{*}{\small 2,322} & \multirow{2}{*}{\begin{tabular}[c]{@{}l@{}}\scriptsize \citet{blodgett-etal-2020-language},\\ \scriptsize \citet{perez-etal-2022-red}\end{tabular}} \\
 &  &  &  &  &  \\ \midrule
\multirow{2}{*}{\small Representation Learning} & \multirow{2}{*}{\small 6,370} & \multirow{2}{*}{\begin{tabular}[c]{@{}l@{}}\scriptsize \citet{reimers-gurevych-2019-sentence},\\ \scriptsize \citet{gao-etal-2021-simcse}\end{tabular}} & \multirow{2}{*}{\small Question Answering} & \multirow{2}{*}{\small 2,208} & \multirow{2}{*}{\begin{tabular}[c]{@{}l@{}}\scriptsize \citet{karpukhin-etal-2020-dense},\\ \scriptsize \citet{liu-etal-2022-challenges})\end{tabular}} \\
 &  &  &  &  &  \\ \midrule
\multirow{2}{*}{\small Text Classification} & \multirow{2}{*}{\small 6,117} & \multirow{2}{*}{\begin{tabular}[c]{@{}l@{}}\scriptsize \citet{wei-zou-2019-eda},\\ \scriptsize \citet{hu-etal-2022-knowledgeable}\end{tabular}} & \multirow{2}{*}{\small Tagging} & \multirow{2}{*}{\small 1,968} & \multirow{2}{*}{\begin{tabular}[c]{@{}l@{}}\scriptsize  \citet{malmi-etal-2019-encode},\\ \scriptsize \citet{wei-etal-2020-novel}\end{tabular}} \\
 &  &  &  &  &  \\ \midrule
\multirow{2}{*}{\small Low-Resource NLP} & \multirow{2}{*}{\small 5,863} & \multirow{2}{*}{\begin{tabular}[c]{@{}l@{}}\scriptsize \citet{gao-etal-2021-making},\\ \scriptsize \citet{liu-etal-2022-makes}\end{tabular}} & \multirow{2}{*}{\small Summarization} & \multirow{2}{*}{\small 1,856} & \multirow{2}{*}{\begin{tabular}[c]{@{}l@{}}\scriptsize \citet{liu-lapata-2019-text},\\ \scriptsize \citet{he-etal-2022-ctrlsum}\end{tabular}} \\
 &  &  &  &  &  \\ \midrule
\multirow{2}{*}{\begin{tabular}[c]{@{}l@{}}\small Dialogue Systems \& \\ \small Conversational Agents\end{tabular}} & \multirow{2}{*}{\small 4,678} & \multirow{2}{*}{\begin{tabular}[c]{@{}l@{}}\scriptsize \citet{zhang-etal-2020-dialogpt},\\ \scriptsize \citet{roller-etal-2021-recipes}\end{tabular}} & \multirow{2}{*}{\small Green \& Sustainable NLP} & \multirow{2}{*}{\small 1,780} & \multirow{2}{*}{\begin{tabular}[c]{@{}l@{}}\scriptsize \citet{strubell-etal-2019-energy},\\ \scriptsize \citet{ben-zaken-etal-2022-bitfit}\end{tabular}} \\
 &  &  &  &  &  \\ \midrule
\multirow{2}{*}{\small Syntactic Parsing} & \multirow{2}{*}{\small 4,028} & \multirow{2}{*}{\begin{tabular}[c]{@{}l@{}}\scriptsize \citet{zhou-zhao-2019-head},\\ \scriptsize \citet{glavas-vulic-2021-supervised}\end{tabular}} & \multirow{2}{*}{\small Cross-Lingual Transfer} & \multirow{2}{*}{\small 1,749} & \multirow{2}{*}{\begin{tabular}[c]{@{}l@{}}\scriptsize \citet{conneau-etal-2020-unsupervised},\\ \scriptsize \citet{feng-etal-2022-language}\end{tabular}} \\
 &  &  &  &  &  \\ \midrule
\multirow{2}{*}{\small Speech \& Audio in NLP} & \multirow{2}{*}{\small 3,915} & \multirow{2}{*}{\begin{tabular}[c]{@{}l@{}}\scriptsize \citet{baevski2022unsupervised},\\ \scriptsize \citet{wang-etal-2020-fairseq}\end{tabular}} & \multirow{2}{*}{\small Morphology} & \multirow{2}{*}{\small 1,749} & \multirow{2}{*}{\begin{tabular}[c]{@{}l@{}}\scriptsize \citet{mccarthy-etal-2020-unimorph},\\ \scriptsize \citet{goldman-etal-2022-un}\end{tabular}} \\
 &  &  &  &  &  \\ \midrule
\multirow{2}{*}{\small Knowledge Representation} & \multirow{2}{*}{\small 2,967} & \multirow{2}{*}{\begin{tabular}[c]{@{}l@{}}\scriptsize \citet{schneider-etal-2022-decade},\\ \scriptsize \citet{safavi-koutra-2021-relational}\end{tabular}} & \multirow{2}{*}{\begin{tabular}[c]{@{}l@{}}\small Explainability \&\\ \small Interpretability in NLP\end{tabular}} & \multirow{2}{*}{\small 1,671} & \multirow{2}{*}{\begin{tabular}[c]{@{}l@{}}\scriptsize \citet{danilevsky-etal-2020-survey},\\ \scriptsize \citet{pruthi-etal-2022-evaluating}\end{tabular}} \\
 &  &  &  &  &  \\ \midrule
\multirow{2}{*}{\small Structured Data in NLP} & \multirow{2}{*}{\small 2,803} & \multirow{2}{*}{\begin{tabular}[c]{@{}l@{}}\scriptsize \citet{herzig-etal-2020-tapas},\\ \scriptsize \citet{yin-etal-2020-tabert}\end{tabular}} & \multirow{2}{*}{\small Robustness in NLP} & \multirow{2}{*}{\small 1,621} & \multirow{2}{*}{\begin{tabular}[c]{@{}l@{}}\scriptsize \citet{hendrycks-etal-2020-pretrained},\\ \scriptsize \citet{meade-etal-2022-empirical}\end{tabular}} \\
 &  &  &  &  &  \\ \bottomrule
\end{tabular}
\renewcommand{\arraystretch}{1}
\caption{Overview of the most popular \acp{fos} in \ac{nlp} literature. Representative papers consist of either highly cited studies or comprehensive surveys on the respective \acp{fos}.}
\label{tab:fos-overview}
\end{table*}

\subsection{Taxonomy of \acp{fos} in \ac{nlp} research (RQ1)}
\label{subsec:rq1}

To develop the taxonomy of \acp{fos} in \ac{nlp} shown in Figure \ref{fig:nlp-taxonomy}, we first examined the submission topics of recent years as listed on the websites of major \ac{nlp} conferences such as ACL, EMNLP, COLING, or IJCNLP. In addition, we reviewed the topics of workshops included in the ACL Anthology to derive further \acp{fos}. In order to include smaller topics that are not necessarily mentioned on conference or workshop websites, we manually reviewed all papers from the recently published EMNLP 2022 Proceedings, extracted their \acp{fos}, and annotated all 828 papers accordingly. This provided us with an initial set of \acp{fos}, which we used to create the first version of the \ac{nlp} taxonomy. 

Based on our initial taxonomy, we conducted semi-structured expert interviews with \ac{nlp} researchers to evaluate and adjust the taxonomy. In the interviews, we placed particular emphasis on the evaluation of the mapping of lower-level \acp{fos} to their higher-level \acp{fos} and the correctness and completeness of \acp{fos} in the \ac{nlp} domain. In total, we conducted more than 20 one-on-one interviews with different domain experts. After conducting the interviews, we noticed that experts demonstrated a high degree of agreement on certain aspects of evaluation, while opinions were highly divergent on other aspects. While we easily implemented changes resulting from high expert agreement, we acted as the final authority in deciding whether to implement a particular change for aspects with low expert agreement. For example, one of the aspects with the highest agreement was that certain lower-level \acp{fos} must be assigned not only to one but also to multiple higher-level \acp{fos}. 

Based on the interview results, we subsequently adjusted the annotations of the 828 EMNLP 2022 papers and developed the final \ac{nlp}-taxonomy, as shown in Figure \ref{fig:nlp-taxonomy}.

\subsection{Field of Study Classification (RQ2)}
\label{subsec:rq2}

We trained a weakly supervised classifier to classify ACL Anthology papers according to the \ac{nlp} taxonomy. To obtain a training dataset, we first defined keywords for each \ac{fos} included in the final taxonomy to perform a database search for relevant articles. Based on the keywords, we created search strings to query the Scopus and arXiv databases. The search string was applied to titles and author keywords, if available. While we limited the Scopus search results to the NLP domain with additional restrictive keywords such as "NLP", "natural language processing", or "computational linguistics", we limited the search in arXiv to the cs.CL domain. We subsequently merged duplicate articles to create a multi-label dataset and removed articles included in the EMNLP 2022 proceedings, as this dataset is used as test set. Finally, we applied a fuzzy string matching heuristic and added missing classes based on the previously defined \ac{fos} keywords that appear twice or more in the article titles or abstracts. The final training dataset consists of 178,521 articles annotated on average with 3.65 different \acp{fos}. On average, each class includes 7936.50 articles, while the most frequent class is represented by 63728 articles and the least frequent class by 141 articles. We split this unevenly distributed dataset into three different random 90/10 training/validation sets and used the human-annotated EMNLP 2022 articles as the test dataset. For multi-label classification, we fine-tuned and evaluated different base models. Training and evaluation details are shown in Appendix \ref{sec:fos-cls-eval-appendix}. We found that SPECTER 2.0 performed best on validation and test data, with average $F_1$ scores of 96.06 and 93.21, respectively, on multiple training runs. Therefore, we selected SPECTER 2.0 as our final classification model, which we subsequently trained on the combined training, validation, and test data. 

Using the final model, we classified all papers included in the ACL Anthology from 1952 to 2022. To obtain our final dataset for analysis, we removed the articles that were not truly research articles, such as prefaces; articles that were not written in English; and articles where the classifier was uncertain and simply predicted every class possible. This final classified dataset includes a total of 74,279 research papers. Table \ref{tab:fos-overview} shows the final classification results with respect to the number of publications for each of the most popular \acp{fos}.

\begin{figure*}[ht!]
    \centering
    \includegraphics[width=\textwidth]{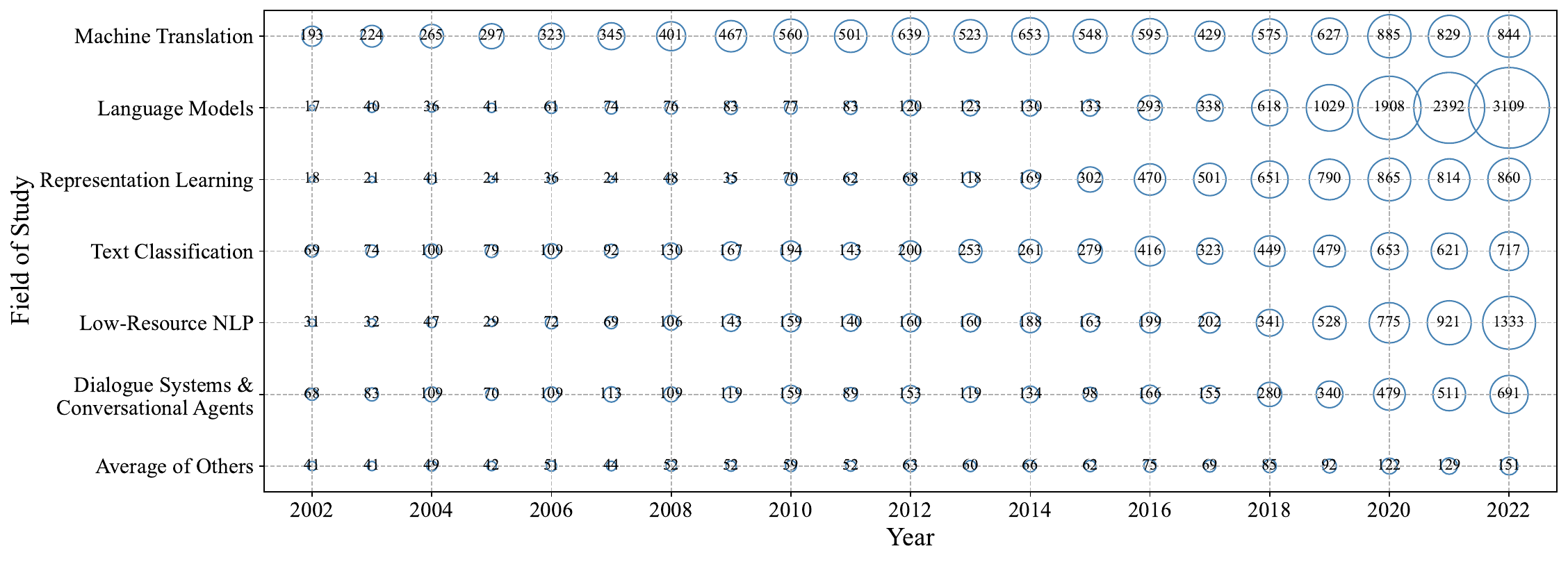}
    \caption{Distribution of number of papers by most popular \acp{fos} from 2002 to 2022.}
    \label{fig:nlp-development}
\end{figure*}

\subsection{Characteristics and Developments of the Research Landscape (RQ3)}
\label{subsec:rq3}
Considering the literature on \ac{nlp}, we start our analysis with the number of studies as an indicator of research interest. The distribution of publications over the 50-year observation period is shown in Figure \ref{fig:acl-anthology-number-of-papers}. While the first publications appeared in 1952, the number of annual publications grew slowly until 2000. Accordingly, between 2000 and 2017, the number of publications roughly quadrupled, whereas in the subsequent five years, it doubled again. We therefore observe a near-exponential growth in the number of \ac{nlp} studies, indicating increasing attention from the research community.

Examining Table \ref{tab:fos-overview} and Figure \ref{fig:nlp-development}, the most popular \acp{fos} in the \ac{nlp} literature and their recent development over time are revealed. While the majority of studies in \ac{nlp} are related to machine translation or language models, the developments of both \acp{fos} are different. Machine translation is a thoroughly researched field that has been established for a long time and has experienced a modest growth rate over the last 20 years. Language models have also been researched for a long time. However, the number of publications on this topic has only experienced significant growth since 2018. Similar differences can be observed when looking at the other popular \acp{fos}. Representation learning and text classification, while generally widely researched, are partially stagnant in their growth. In contrast, dialogue systems \& conversational agents and particularly low-resource \ac{nlp} continue to exhibit high growth rates in the number of studies. Based on the development of the average number of studies on the remaining \acp{fos} in Figure \ref{fig:nlp-development}, we observe a slightly positive growth overall. However, the majority of \acp{fos} are significantly less researched than the most popular \acp{fos}. We conclude that the distribution of research across \acp{fos} is extremely unbalanced and that the development of \ac{nlp} research is largely shaped by advances in a few highly popular \acp{fos}. 

\begin{figure*}[ht!]
    \centering
    \includegraphics[width=\textwidth]{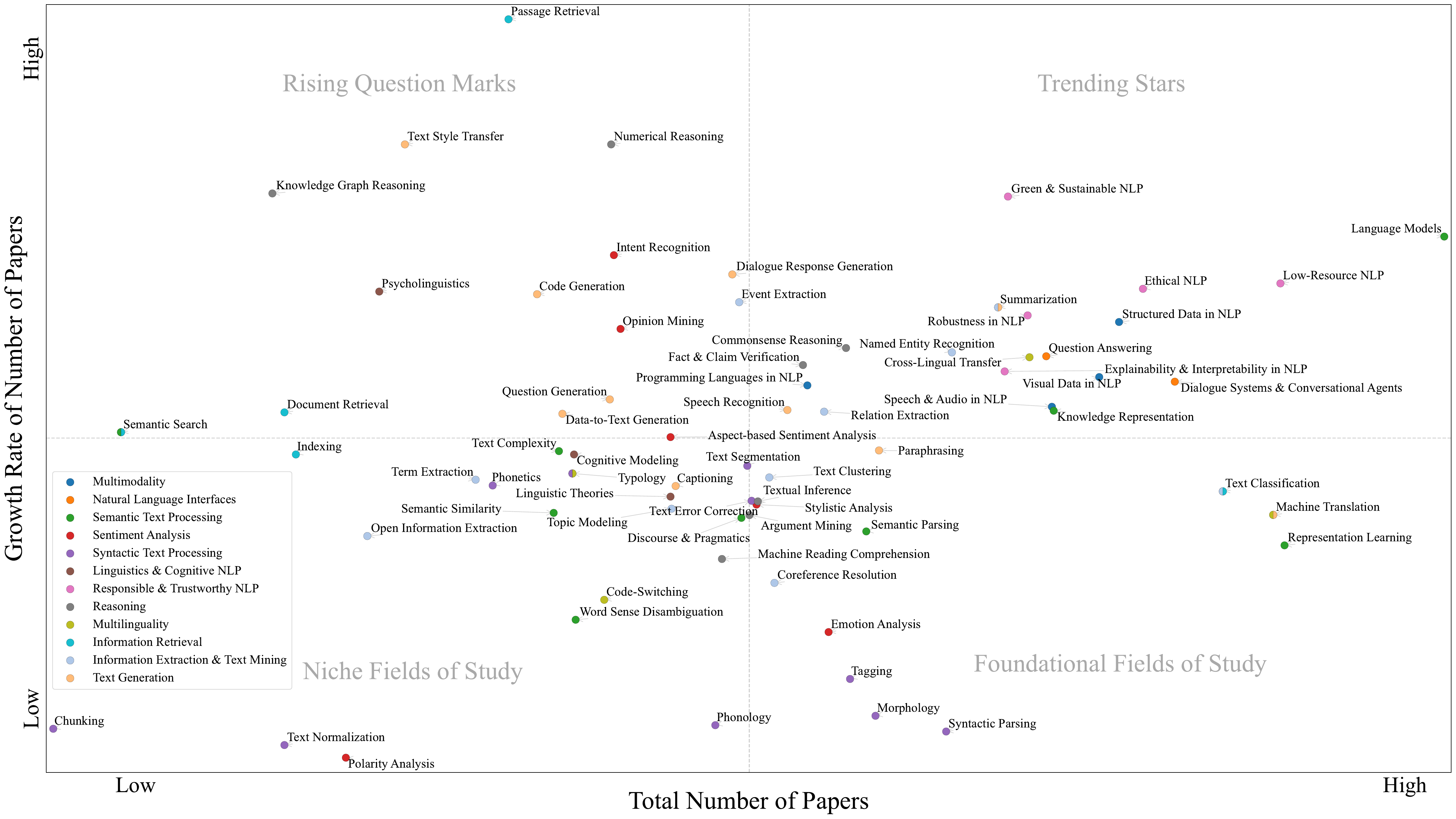}
    \caption{Growth-share matrix of \acp{fos} in \ac{nlp}. The growth rates and total number of works for each \ac{fos} are calculated from the start of 2018 to the end of 2022. To obtain a more uniform distribution of the data, we apply the Yeo-Johnson transformation \cite{10.1093/biomet/87.4.954}.}
    \label{fig:nlp-trends}
\end{figure*}

\subsection{Research Trends and Directions for Future Work (RQ4)}
\label{subsec:rq4}

Figure \ref{fig:nlp-trends} shows the growth-share matrix of \acp{fos} in \ac{nlp} research inspired by \citet{henderson_1970}. We use it to examine current research trends and possible future research directions by analyzing the growth rates and total number of papers related to the various \acp{fos} in \ac{nlp} between 2018 and 2022. The upper right section of the matrix consists of \acp{fos} that exhibit a high growth rate and simultaneously a large number of papers overall. Given the growing popularity of \acp{fos} in this section, we categorize them as \textit{trending stars}. The lower right section contains \acp{fos} that are very popular but exhibit a low growth rate. Usually, these are \acp{fos} that are essential for \ac{nlp} research but already relatively mature. Hence, we categorize them as \textit{foundational \acp{fos}}. The upper left section of the matrix contains \acp{fos} that exhibit a high growth rate but only very few papers overall. Since the progress of these \acp{fos} is rather promising, but the small number of overall papers renders it difficult to predict their further developments, we categorize them as \textit{rising question marks}. The \acp{fos} in the lower left of the matrix are categorized as \textit{niche \acp{fos}} owing to their low total number of papers and their low growth rates.

Figure \ref{fig:nlp-trends} shows that language models are currently receiving the most attention, which is also consistent with the observations from Table \ref{tab:fos-overview} and Figure \ref{fig:nlp-development}. Based on the latest developments in this area, this trend is likely to continue and accelerate in the near future. Text classification, machine translation, and representation learning rank among the most popular \acp{fos} but only show marginal growth. In the long term, they may be replaced by faster-growing fields as the most popular \acp{fos}. 

In general, \acp{fos} related to syntactic text processing exhibit negligible growth and low popularity overall. Conversely, \acp{fos} concerned with responsible \& trustworthy \ac{nlp}, such as green \& sustainable \ac{nlp}, low-resource \ac{nlp}, and ethical \ac{nlp} tend to exhibit a high growth rate and also high popularity overall. This trend can also be observed in the case of structured data in \ac{nlp}, visual data in \ac{nlp}, and speech \& audio in \ac{nlp}, all of which are concerned with multimodality. In addition, natural language interfaces involving dialogue systems \& conversational agents, and question answering are becoming increasingly important in the research community. We conclude that in addition to language models, responsible \& trustworthy \ac{nlp}, multimodality, and natural language interfaces are likely to characterize the \ac{nlp} research landscape in the near future.  

Further notable developments can be observed in the area of reasoning, specifically with respect to knowledge graph reasoning and numerical reasoning and in various \acp{fos} related to text generation. Although these \acp{fos} are currently still relatively small, they apparently attract more and more interest from the research community and show a clear positive tendency toward growth. 

\begin{figure*}[ht!]
    \centering
    \includegraphics[width=\textwidth]{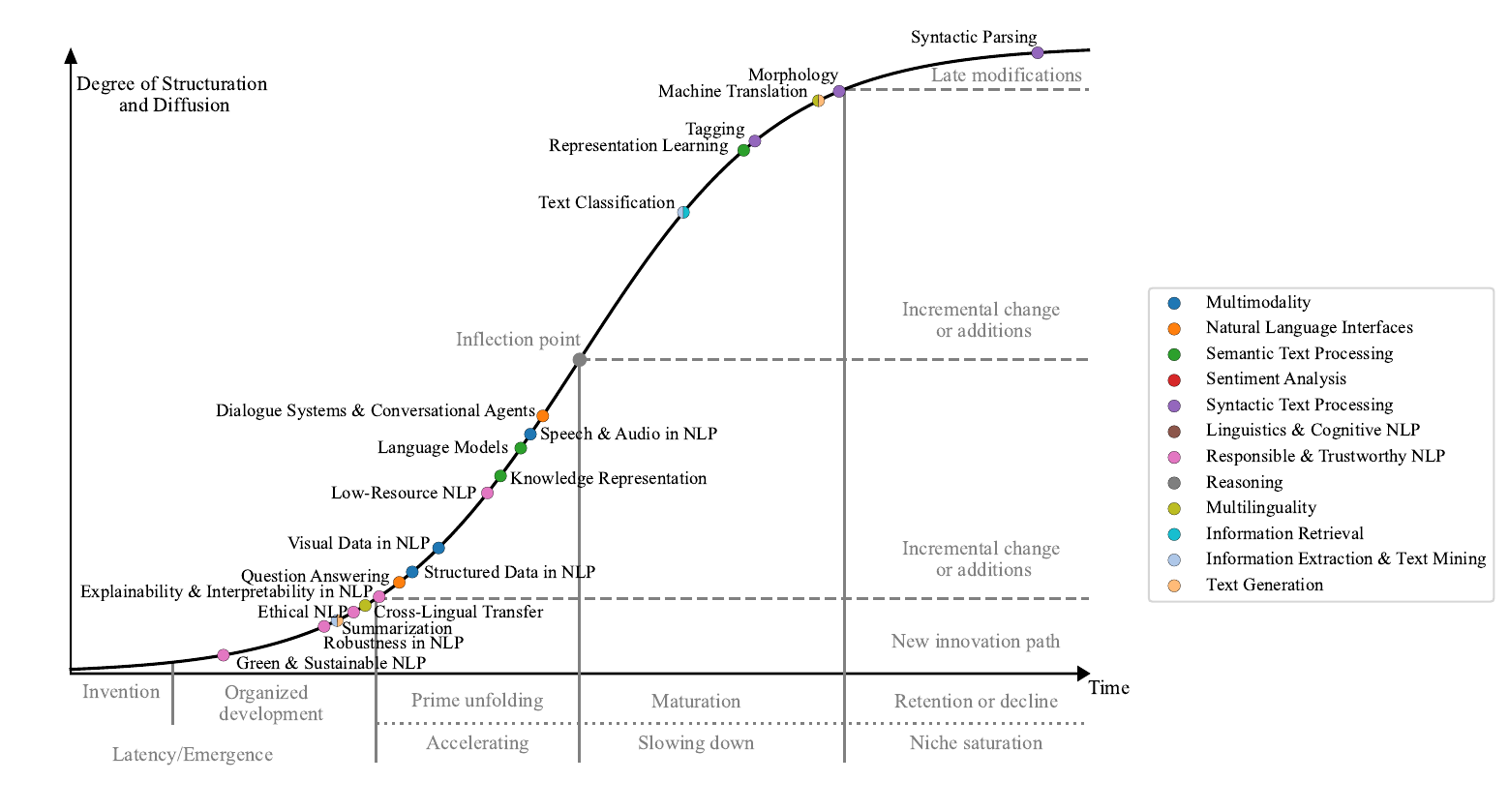}
    \caption{Innovation life cycle of the most popular \acp{fos} in \ac{nlp}. \acp{fos} on the left side of the curve are at the beginning of their life cycle. They have just been invented or are in an early phase, where innovation on \acp{fos} accelerates by a rising number of studies. After passing the inflection point, the \acp{fos} move towards the end of their innovation life cycle, where research on \acp{fos} is retained or declines and only late modifications are added to the \acp{fos}.}
    \label{fig:innovation-analysis}
\end{figure*}

Figure \ref{fig:innovation-analysis} shows the innovation life cycle of the most popular \acp{fos} in \ac{nlp} adapted from the \textit{diffusion of innovations} theory \cite{rogers1962diffusion} and inspired by \citet{Huber_2005}. The central assumption of the innovation life cycle theory is that for each innovation (or in this case \ac{fos}), the number of published research per year is normally distributed over time, while the total number of published research reaches saturation according to a sigmoid curve. Appendix \ref{sec:innovation-life-cycle-appendix} shows how the positions of \acp{fos} on the innovation life cycle curve are determined.

From Figure \ref{fig:innovation-analysis}, we observe that \acp{fos} related to syntactic text processing are already relatively mature and approaching the end of the innovation life cycle. Particularly, syntactic parsing is getting near the end of its life cycle, with only late modifications being researched. While Table \ref{tab:fos-overview} shows that machine translation, representation learning, and text classification are very popular overall, Figure \ref{fig:innovation-analysis} reveals that they have passed the inflection point of the innovation life cycle curve and their development is currently slowing down. They are adopted by most researchers but show stagnant or negative growth, as also indicated in Figure \ref{fig:nlp-trends}. However, most of \acp{fos} have not yet reached the inflection point and are still experiencing increasing growth rates, while research on these \acp{fos} is accelerating. Especially \acp{fos} related to responsible \& trustworthy \ac{nlp}, multimodality, and natural language interfaces are just beginning their innovation life cycle, suggesting that research in these areas will likely accelerate in the following years. This is also in line with the observations from Figure \ref{fig:nlp-trends}, where most of the \acp{fos} related to these areas are categorized as \textit{trending stars}. Further, we observe that language models have passed the first two stages of innovation and are currently in their prime unfolding phase. They are adopted by a large number of researchers and research on them is still accelerating. Comparing this to Figure \ref{fig:nlp-trends}, where language models are among the most trending \acp{fos}, we conclude that this trend is likely to continue in the near future and is unlikely to slow down anytime soon. 

\section{Discussion}
\label{sec:discussion}

The observations of our comprehensive study reveal several insights that we can situate to related work. Since the first publications in 1952, researchers have paid increasing attention to the field of \ac{nlp}, particularly after the introduction of Word2Vec \cite{mikolov2013efficient} and accelerated by BERT \cite{devlin-etal-2019-bert}. This observed growth in research interest is in line with the study of \citet{mohammad-2020-nlp-scholar}. Historically, machine translation was one of the first research fields in \ac{nlp} \cite{jones1994natural}, which continues to be popular and steadily growing nowadays. However, recent advances in language model training have sparked increasing research efforts in this field, as shown in Figure \ref{fig:nlp-development}. Since scaling up language models significantly enhance performance on downstream tasks \cite{NEURIPS2020_1457c0d6,DBLP:journals/corr/abs-2001-08361,wei2022emergent,hoffmann2022an}, researchers continue to introduce increasingly larger language models \cite{HAN2021225}. However, training and using these large language models involves significant challenges, including computational costs \cite{DBLP:journals/corr/abs-2104-04473}, environmental issues \cite{strubell-etal-2019-energy}, and ethical considerations \cite{perez-etal-2022-red}. As a result, a recent increase in research efforts has been noted to render language models and \ac{nlp} more responsible \& trustworthy in general, as shown in Figure \ref{fig:nlp-trends} and Figure \ref{fig:innovation-analysis}. Additionally, recent advances aim to train large-scale multimodal language models capable of understanding and generating natural language text and performing all types of downstream tasks while interacting with humans through natural language input prompts \cite{openai2023gpt4}. From our observations in Figure \ref{fig:nlp-trends} and Figure \ref{fig:innovation-analysis}, we again find support for this trend in \ac{nlp} literature for multimodality, text generation, and natural language interfaces. 

Although language models have achieved remarkable success on various \ac{nlp} tasks, their inability to reason is often seen as a limitation that cannot be overcome by increasing the model size alone \cite{rae2022scaling,wei2022chain,wang2023selfconsistency}. Although reasoning capabilities are a crucial prerequisite for the reliability of language models, this field is still relatively less researched and receives negligible attention. While Figure \ref{fig:nlp-trends} exhibits high growth rates for knowledge graph reasoning and numerical reasoning in particular, research related to reasoning is still rather underrepresented compared to the more popular \acp{fos}.

\section{Conclusion}
\label{sec:conclusion}

Recent years have witnessed an increasing prominence of \ac{nlp} research. To summarize recent developments and provide an overview of this research area, we defined a taxonomy of \acp{fos} in \ac{nlp} and analyzed recent research developments.

Our findings show that a large number of \acp{fos} have been studied, including trending fields such as multimodality, responsible \& trustworthy \ac{nlp}, and natural language interfaces. While recent developments are largely a result of recent advances in language models, we have noted a lack of research pertaining to teaching these language models to reason and thereby afford more reliable predictions. 

\section{Limitations}
\label{sec:limitationsf}

Constructing the taxonomy highly depends on the personal decisions of the authors, which can bias the final result. The taxonomy may not cover all possible \acp{fos} and offers potential for discussions, as domain experts have inherently different opinions. As a countermeasure, we aligned the opinions of multiple domain experts and designed the taxonomy at a higher level, allowing non-included \acp{fos} to be considered as possible subtopics of existing ones. 

For this study, we limited our analysis to papers published in the ACL Anthology, which typically feature research presented at major international conferences and are written in English. However, research communities that publish their work in regional venues exist, often in languages other than English. In addition, \ac{nlp} research is also presented at other prominent global conferences such as AAAI, NeurIPS, ICLR, or ICML. Therefore, the findings we report in this study pertain specifically to \ac{nlp} research presented at major international conferences and journals in English.

Furthermore, the accuracy of the classification results poses another threat to the validity of our study. Data extraction bias and classification model errors may negatively affect the results. To mitigate this risk, the authors regularly discussed the used classification schemes and conducted a thorough evaluation of the performance of the classification model.

\section*{Acknowledgments}

We would like to thank Phillip Schneider, Stephen Meisenbacher, Mahdi Dhaini, Juraj Vladika, Oliver Wardas, Anum Afzal, Wessel Poelman, and Alexander Blatzheim of sebis for helpful discussions and valuable feedback.

\bibliography{anthology,custom}
\bibliographystyle{acl_natbib}

\clearpage
\appendix

\section{Appendix}
\label{sec:appendix-a}

\subsection{Fields of Study Descriptions}\label{sec:fos-descriptions}

In the following, we explain the fields of study concepts included in the \ac{nlp} taxonomy in Figure \ref{fig:nlp-taxonomy}.

\subsubsection{Multimodality}
\label{subsubsec:multimodality}
Multimodality refers to the capability of a system or method to process input of different types or “modalities” \cite{garg-etal-2022-multimodality}. We distinguish between systems that can process text in natural language along with \textbf{visual data}, \textbf{speech \& audio}, \textbf{programming languages}, or \textbf{structured data} such as tables or graphs. 

\subsubsection{Natural Language Interfaces}
\label{subsubsec:natural-language-interfaces}
Natural language interfaces can process data based on natural language queries \cite{voigt-etal-2021-challenges}, usually implemented as \textbf{question answering} or \textbf{dialogue \& conversational systems}.

\subsubsection{Semantic Text Processing}
\label{subsubsec:semantic-text-processing}
This high-level \ac{fos} includes all types of concepts that attempt to derive meaning from natural language and enable machines to interpret textual data semantically. One of the most powerful \ac{fos} in this regard are \textbf{language models} that attempt to learn the joint probability function of sequences of words \cite{NIPS2000_728f206c}. Recent advances in language model training have enabled these models to successfully perform various downstream \ac{nlp} tasks \cite{soni-etal-2022-human}. In \textbf{representation learning}, semantic text representations are usually learned in the form of embeddings \cite{fu-etal-2022-contextual,schopf2023efficient,schopf2023aspectcse}, which can be used to compare the \textbf{semantic similarity} of texts in \textbf{semantic search} settings \cite{reimers-gurevych-2019-sentence}. Additionally, \textbf{knowledge representations}, e.g., in the form of knowledge graphs, can be incorporated to improve various \ac{nlp} tasks \cite{schneider-etal-2022-decade}. 

\subsubsection{Sentiment Analysis}
\label{subsubsec:sentiment-analysis}
Sentiment analysis attempts to identify and extract subjective information from texts \cite{Wankhade2022}. Usually, studies focus on extracting \textbf{opinions}, \textbf{emotions}, or \textbf{polarity} from texts. More recently, \textbf{aspect-based sentiment analysis} emerged as a way to provide more detailed information than general sentiment analysis, as it aims to predict the sentiment polarities of given aspects or entities in text \cite{xue-li-2018-aspect}.

\subsubsection{Syntactic Text Processing}
\label{subsubsec:syntactic-text-processing}
This high-level \ac{fos} aims at analyzing the grammatical syntax and vocabulary of texts \cite{7991677}. Representative tasks in this context are \textbf{syntactic parsing} of word dependencies in sentences, \textbf{tagging} of words to their respective part-of-speech, \textbf{segmentation} of texts into coherent sections, or \textbf{correction of erroneous texts} with respect to grammar and spelling. 

\subsubsection{Linguistics \& Cognitive NLP}
\label{subsubsec:linguistics-cognitive-nlp}
Linguistics \& Cognitive \ac{nlp} deals with natural language based on the assumptions that our linguistic abilities are firmly rooted in our cognitive abilities, that meaning is essentially conceptualization, and that grammar is shaped by usage \cite{cognitive-linguistics}. Many different \textbf{linguistic theories} are present that generally argue that language acquisition is governed by universal grammatical rules that are common to all typically developing humans \cite{WISE2017}. \textbf{Psycholinguistics} attempts to model how a human brain acquires and produces language, processes it, comprehends it, and provides feedback \cite{Balamurugan2018IntroductionTP}. \textbf{Cognitive modeling} is concerned with modeling and simulating human cognitive processes in various forms, particularly in a computational or mathematical form \cite{Sun2020}.

\subsubsection{Responsible \& Trustworthy NLP}
\label{subsubsec:responsible-nlp}
Responsible \& trustworthy \ac{nlp} is concerned with implementing methods that focus on fairness, \textbf{explainability}, accountability, and \textbf{ethical} aspects at its core \cite{BARREDOARRIETA202082}. \textbf{Green \& sustainable \ac{nlp}} is mainly focused on efficient approaches for text processing, while \textbf{low-resource \ac{nlp}} aims to perform \ac{nlp} tasks when data is scarce. Additionally, \textbf{robustness in \ac{nlp}} attempts to develop models that are insensitive to biases, resistant to data perturbations, and reliable for out-of-distribution predictions.

\begin{table*}[!b]
    \centering
    \resizebox{1.94\columnwidth}{!}{%
    \renewcommand{\arraystretch}{1.2} %
   
    \begin{adjustbox}{center}

    \begin{tabular}{lc ccc ccc}
    \toprule
    \multicolumn{2}{l}{\textbf{Dataset $\rightarrow$}} & \multicolumn{3}{c}{\textit{Validation}} & \multicolumn{3}{c}{\textit{Test}} \\

    \cmidrule(lr){3-5}
    \cmidrule(lr){6-8}
    
    \multicolumn{2}{l}{\textbf{Model $\downarrow$}} &      \textbf{P} &      \textbf{R} &    $\textbf{F}_{1}$ &            \textbf{P} &      \textbf{R} &    $\textbf{F}_{1}$  \\
    
    \midrule

    \multicolumn{2}{l}{BERT} &
    \textbf{96.57\small{$\pm$0.14}} &  95.43\small{$\pm$0.16} & 96.00\small{$\pm$0.03} & 89.77\small{$\pm$0.20} & 93.58\small{$\pm$0.07} & 91.64\small{$\pm$0.10} \\
    \multicolumn{2}{l}{RoBERTa} &
    95.77\small{$\pm$0.19} & 95.19\small{$\pm$0.16} & 95.48\small{$\pm$0.17} & 87.46\small{$\pm$2.75} & 93.29\small{$\pm$0.10} & 90.27\small{$\pm$1.42} \\
    \multicolumn{2}{l}{SciBERT} & 
    96.44\small{$\pm$0.17} & 95.65\small{$\pm$0.14} & 96.05\small{$\pm$0.10} & 90.18\small{$\pm$3.17} & \textbf{94.05\small{$\pm$0.06}} & 92.06\small{$\pm$1.65} \\
    \multicolumn{2}{l}{SPECTER 2.0} & 
    96.44\small{$\pm$0.11} & 95.69\small{$\pm$0.14} & \textbf{96.06\small{$\pm$0.08}} & \textbf{92.46\small{$\pm$2.58}} & 93.99\small{$\pm$0.22} & \textbf{93.21\small{$\pm$1.39}} \\
    \multicolumn{2}{l}{SciNCL} & 
    96.39\small{$\pm$0.11} & \textbf{95.71\small{$\pm$0.09}} & 96.05\small{$\pm$0.04} & 89.97\small{$\pm$1.85} & 93.74\small{$\pm$0.18} & 91.81\small{$\pm$0.93} \\
  
    \bottomrule
    \end{tabular}
\end{adjustbox}%
}
\caption{Evaluation results for classifying papers according to the \ac{nlp} taxonomy on three runs over different random train/validation splits. Since the distribution of classes is very unbalanced, we report micro scores.}
\label{tab:cls-evaluation}
\end{table*}

\subsubsection{Reasoning}
\label{subsubsec:reasoning}
Reasoning enables machines to draw logical conclusions and derive new knowledge based on the information available to them, using techniques such as deduction and induction. \textbf{Argument mining} automatically identifies and extracts the structure of inference and reasoning expressed as arguments presented in natural language texts \cite{lawrence-reed-2019-argument}. \textbf{Textual inference}, usually modeled as entailment problem, automatically determines whether a natural-language \textit{hypothesis} can be inferred from a given \textit{premise} \cite{maccartney-manning-2007-natural}. \textbf{Commonsense reasoning} bridges premises and hypotheses using world knowledge that is not explicitly provided in the text \cite{ponti-etal-2020-xcopa}, while \textbf{numerical reasoning} performs arithmetic operations \cite{al-negheimish-etal-2021-numerical}. \textbf{Machine reading comprehension} aims to teach machines to determine the correct answers to questions based on a given passage \cite{Zhang_Yang_Zhao_2021}.

\subsubsection{Multilinguality}
\label{subsubsec:multilinguality}
Multilinguality tackles all types of \ac{nlp} tasks that involve more than one natural language and is conventionally studied in \textbf{machine translation}. Additionally, \textbf{code-switching} freely interchanges multiple languages within a single sentence or between sentences \cite{diwan21_interspeech}, while \textbf{cross-lingual transfer} techniques use data and models available for one language to solve \ac{nlp} tasks in another language. 

\subsubsection{Information Retrieval}
\label{subsubsec:information-retrieval}
Information retrieval is concerned with finding texts that satisfy an information need from within large collections \cite{manning_raghavan_schütze_2008}. Typically, this involves retrieving \textbf{documents} or \textbf{passages}.

\subsubsection{Information Extraction \& Text Mining}
\label{subsubsec:information-extraction}
This \ac{fos} focuses on extracting structured knowledge from unstructured text and enables the analysis and identification of patterns or correlations in data \cite{bdcc4010001}. \textbf{Text classification} automatically categorizes texts into predefined classes \citet{schopf_etal_webist_21,10.1145/3582768.3582795,10.1007/978-3-031-24197-0_4}, while \textbf{topic modeling} aims to discover latent topics in document collections \cite{Grootendorst2022BERTopicNT}, often using \textbf{text clustering} techniques that organize semantically similar texts into the same clusters. \textbf{Summarization} produces summaries of texts that include the key points of the input in less space and to keep repetition to a minimum \noindent \cite{ELKASSAS2021113679}. Additionally, the information extraction \& text mining \ac{fos} also includes \textbf{named entity recognition}, which deals with the identification and categorization of named entities \cite{leitner-etal-2020-dataset}, \textbf{coreference resolution} that aims to identify all references to the same entity in discourse \cite{yin-etal-2021-signed}, \textbf{term extraction} that aims to extract relevant terms such as keywords or keyphrases \cite{rigouts-terryn-etal-2020-termeval,schopf_etal_kdir_22}, \textbf{relation extraction} that aims to extract relations between entities, and \textbf{open information extraction} that facilitates the domain-independent discovery of relational tuples \cite{yates-etal-2007-textrunner}.

\subsubsection{Text Generation}
\label{subsubsec:text generation}
The objective of text generation approaches is to generate texts that are both comprehensible to humans and indistinguishable from text authored by humans. Accordingly, the input usually consists of text, such as in \textbf{paraphrasing} that renders the text input in a different surface form while preserving the semantics \cite{niu-etal-2021-unsupervised}, \textbf{question generation} that aims to generate a fluid and relevant question given a passage and a target answer \cite{song-etal-2018-leveraging}, or \textbf{dialogue-response generation} which aims to generate natural-looking text relevant to the prompt \cite{zhang-etal-2020-dialogpt}. In many cases, however, the text is generated as a result of input from other modalities, such as in the case of \textbf{data-to-text generation} that generates text based on structured data such as tables or graphs \cite{kale-rastogi-2020-text}, \textbf{captioning} of images or videos, or \textbf{speech recognition} that transcribes a speech waveform into text \cite{baevski2022unsupervised}.

\subsection{Evaluating Fields of Study Classification Models}\label{sec:fos-cls-eval-appendix}
For multi-label classification, BERT \cite{devlin-etal-2019-bert}, RoBERTa \cite{liu2019roberta}, SciBERT \cite{beltagy-etal-2019-scibert}, SPECTER 2.0 \cite{cohan-etal-2020-specter,Singh2022SciRepEvalAM}, and SciNCL \cite{ostendorff-etal-2022-neighborhood} models were fine-tuned in their base versions on the three different training datasets and evaluated on their respective validation and test datasets. We trained all models for three epochs, using a batch size of 8, a learning rate of $5e-5$, and the AdamW optimizer \cite{loshchilov2018decoupled}. The evaluation results are shown in Table \ref{tab:cls-evaluation}.

\subsection{Calculating the Positions of Fields of Study on the Innovation Life Cycle Curve}\label{sec:innovation-life-cycle-appendix}
We consider the following aspects which influence the position of a \acp{fos} on the innovation life cycle curve:

\begin{itemize}
    \item A high growth rate in the number of publications indicates that \ac{fos} are at the beginning of their life cycle, while a stagnant or negative growth rate indicates a tendency toward maturation.
    \item If the number of recently published papers accounts for a significant percentage of the total number of papers published on a certain \ac{fos} over time, this indicates the new development of a \ac{fos}.
    \item A high percentage of recent publications on a certain \ac{fos} compared to the total number of recent publications on all \acp{fos} indicates the maturity of a \ac{fos} and its adoption by the majority of researchers. 
\end{itemize}

Accordingly, we define the position of a set of \acp{fos} $F = \{f_{1},..,f_{m}\}$ on the x-axis of the innovation life cycle curve as:
\begin{equation}
\label{eqn:innovation_eqn_x}
    x_{f_{i}} = \log \biggl(\frac{1}{g_{t-n,t}(\scriptstyle f_{i}\displaystyle)} \cdot \frac{1}{h_{t-n,t}(\scriptstyle f_{i}\displaystyle)} \cdot k_{t-n,t}(\scriptstyle f_{i}\displaystyle)\biggl)
\end{equation}
where $t$ is a specific year, ${g_{t-n,t}(\scriptstyle f_{i}\displaystyle)}$ is the growth rate of the number of publications for a particular \ac{fos}, normalized between $1e^{-10}$ and $1$, ${h_{t-n,t}(\scriptstyle f_{i}\displaystyle)}$ is the percentage of the number of papers for a particular \ac{fos} in a chosen time period compared to the total number of papers for the same \ac{fos} over the entire observed time period, and $k_{t-n,t}(\scriptstyle f_{i}\displaystyle)$ is the percentage of the number of publications for a specific \ac{fos} in a chosen time period compared to all publications in the same time period across all \acp{fos}. We choose a five-year time period with $t=2022$ and $n=5$, while the entire observed time period ranges from 1952 to 2022. Finally, to map the \acp{fos} to the innovation life cycle curve, we normalize $X = \{x_{f_{1}},..,x_{f_{m}}\}$ between lower and upper bounds of $-5$ and $5$ as $X' = \{x'_{f_{1}},..,x'_{f_{m}}\}$ and calculate their position on the y-axis of the innovation life cycle curve as:
\begin{equation}
\label{eqn:innovation_eqn_y}
    y_{f_{i}} = \frac{1}{1+e^{-x'_{f_{i}}}}. 
\end{equation} 

\end{document}